%% file: latex/acl_latex.tex
\pdfoutput=1
\documentclass[11pt]{article}

\usepackage{acl}

\usepackage{times}
\usepackage{latexsym}
\usepackage[T1]{fontenc}
\usepackage[utf8]{inputenc}
\usepackage{microtype}
\usepackage{inconsolata}
\usepackage{graphicx}
\usepackage{booktabs}
\usepackage{tabularx}
\usepackage{amsmath}
\usepackage{algorithm}
\usepackage{subcaption}
\usepackage{algpseudocode}
\usepackage{enumitem}
\usepackage{titlesec}
\usepackage{xurl}
\usepackage{longtable}
\usepackage{comment}

\title{TRIM: Token Reduction and Inference Modeling for Cost-Effective Language Generation}

\author{Alfredo Garrachón Ruiz \and Tomás de la Rosa \and Daniel Borrajo \\
        AI Research, JPMorganChase}

\begin{document}
\maketitle
\begin{abstract}
The high inference cost of Large Language Models (LLMs) poses challenges, especially for tasks requiring lengthy outputs. However, natural language often contains redundancy, which presents an opportunity for optimization. We have observed that LLMs can generate distilled language (i.e., concise outputs that retain essential meaning) when prompted appropriately.
We propose TRIM, a pipeline for saving computational cost in which the LLM omits a predefined set of semantically irrelevant and easily inferable words based on the context during inference. Then, a specifically trained smaller language model with lower inference cost reconstructs the distilled answer into the ideal answer. Our experiments show promising results, particularly on the proposed \textit{NaLDA} evaluation dataset focused on the reconstruction task, with 19.4\% saved tokens on average for GPT-4o and only a tiny decrease in evaluation metrics. This suggests that the approach can effectively balance efficiency and accuracy in language processing tasks.
\end{abstract}

\input{Sections/1.introduction}

\input{Sections/2.related_work}

\input{Sections/3.exploratory}

\input{Sections/4.redundant_language}

\input{Sections/5.1.pipeline}

\input{Sections/5.datasets}

\input{Sections/7.evaluation}

\input{Sections/10.discussion}

\input{Sections/11.conclusion}

\input{Sections/13.limitations}

\input{Sections/12.disclaimer}

\bibliography{references}
\input{Sections/14.Supplementary_Material}

\end{document}

%% file: Sections/1.introduction.tex
\section{Introduction}
Large language models (LLMs) have shown remarkable capabilities across a wide range of tasks, from natural language understanding to creative content generation. However, the computational cost of inference and the associated energy consumption present significant challenges. As the demand for AI applications continues to grow, these costs are expected to escalate, raising concerns about sustainability and accessibility \cite{wu2022sustainable}. Addressing these issues is crucial to ensure that future advancements in AI remain feasible and widely deployable.
Hence, we propose a new technique to save tokens at inference time by skipping less informative \textit{function words} that are easily-inferable based on the context. Later on, we propose to regenerate them based on the context with a cost-effective alternative, leveraging the fact that natural language is considerably redundant.


As part of our feasibility analysis, we  verified that state-of-the-art LLMs can omit a predefined set of easily-inferable words, even without being specifically designed for this task.
Studies in cognitive linguistics suggest that human readers naturally infer missing function words based on context \cite{ferreira2007goodenough}. 
Similarly, transformer-based models use masked language modeling to predict the intentionally masked tokens based on the context \cite{devlin2019bertpretrainingdeepbidirectional}. Our hypothesis is that small language models can mirror this human reading abilities of filling the gaps for low-information easily-inferable words.




In this work, the main contribution is the proposal of a novel pipeline called TRIM.  This pipeline reduces the inference cost of LLMs by omitting easily-inferable  words, which are later on reconstructed using much smaller specifically trained language models. This approach is orthogonal to other optimization techniques, and could be applicable as LLMs continue to grow in size and capabilities. To support this pipeline, we propose also an algorithm designed to sort a list of words by how easily they can be inferred from the context based on a corpus of texts, as well as a technique to experimentally define the length of the set of words based on metrics losses as the percentage of saved tokens increases. 
To support our evaluation, we developed a new dataset specifically designed for the reconstruction task and covering five different text generation types. Additionally, we introduce a new evaluation metric tailored to assess reconstruction performance, focusing on the words omitted and subsequently inferred.
The results show that even omitting around one fifth of the full text, models that are hundreds of times smaller can reconstruct distilled text without losing significant text quality according to different metrics, keeping the semantic meaning practically intact.

%% file: Sections/2.related_work.tex
\section{Related Work}
In this section we review several approaches to optimize LLM inference which are related to our work.

\subsection{Prompt Compression}
One alternative to decrease the computational cost is setting the focus on reducing the input by prompt compression. This could be a token-level removal of less relevant words~\cite{jiang2023-llmlingua,pan2024_llmlingua2, fu2024lazyllm}, a learned policy guided by reinforcement learning~\cite{Jung_2024}, context-aware sentence encoding approached~\cite{liskavets2024prompt}, and more~\cite{zhang2025empiricalstudypromptcompression}. 
These techniques operate on the input rather than the output, making our approach complementary to prompt compression. 


\subsection{Inference Optimizations}
The research of general inference optimization techniques around LLM has been a hot topic with several different techniques proposed. Some of them relate to the computation of the tokens from the prompt using parallel and key-value cached techniques~\cite{miao2023_inferserv_survey, patel2024splitwise}. In other implementations, smaller LLM models can be used to generate multiple output sequences in parallel which are then merged and verified~\cite{miao2023specinfer}. Another way to save costs at inference time for complex tasks consists of building a sequence of increasing-size LLMs, where a (larger) model at step $N$ is only queried when the answer of the model at step $N -1$ is not deem reliable~\cite{chen2023cascadingllm}. Furthermore, several techniques around the instance and cluster level optimization have been proposed~\cite{zhen2025tamingtitanssurveyefficient}. All of these techniques are compatible with our approach as they tackle the problem from different points of view, making them orthogonal to our proposal.

\subsubsection{Text Reconstruction}
Text reconstruction has been studied since long time ago. Text with mutilated characters require reconstruction~\cite{miller1957reconstruction} due to errors that occur in message transmission or typing. One of the core learning ideas in Natural Language Processing (NLP) is having a vectorized representation that allows models to reconstruct back the original text~\cite{oshri2015autoenc_reconstruc}. Nevertheless, NLP tasks, such as masking~\cite{devlin2019bertpretrainingdeepbidirectional} or in a more general scope, infilling~\cite{donahue2020infill}, rely on markers indicating where to substitute the text. In our approach, the input text at inference time would not have indication of what positions contained an omitted word.  Text with omitted words can also be used as negative examples, to improve the performance of automatic translators~\cite{yang2019_nmtwordomission}, but to our knowledge omitted-content texts have not been used to train for inference efficiencies purposes.

%% file: Sections/3.exploratory.tex
\section{Exploratory Research}
\label{exploratory}
The feasibility of our approach relies on the LLM being able to perform the task of generating an answer omitting a set of words that do not contribute to the semantic meaning. This is a task for which the LLM was not originally trained. To verify this assumption, we performed an exploratory analysis where we used the GPT-4o model from OpenAI~\cite{openai}, and developed a prompt to carry out this task indicating also that omitted words should be easily inferrable in a later stage. The prompt includes a pre-defined set of words composed by determiners and conjunctions with a total of 23 words such as: (\textit{the, a, an, and, but, so, that}). See supplementary material (Section~\ref{supplementary_exploratory}) for details regarding the full set of words, designed prompt and further analysis.
We took 500 knowledge questions (See dataset in Section~\ref{datasets}) and gave them to the LLM in two rounds, first with the previously defined prompt appended and then without it to collect the original expected answer. We computed the average number of words from the set present in both types of answers (Table~\ref{tab:exploratory}). We observed how the LLM is able to complete the task effectively, generating the answer omitting nearly all words. Interestingly, the average is not zero, indicating that in some scenarios the LLM preserves some of these words pro-actively deciding that they are not easily-inferable a posteriori based on the context in that scenario.

\begin{table}[h!]
\centering
\small
\begin{tabular}{lrr}
\toprule
Text Type &   Avg. count from word set &  Std. Deviation \\
\midrule
Original & 22.50 &               5.77 \\
Distilled &  0.32 &               0.62 \\
\bottomrule
\end{tabular}
\caption{Average count of the words present in the text.}
\label{tab:exploratory}
\end{table}

%% file: Sections/4.redundant_language.tex
\section{Semantic Relevance Of Words}
The importance of certain words in conveying semantic meaning is a well-documented phenomenon in English~\cite{zipf1949human}.
Common words, such as articles and determiners, often contribute little to the overall meaning of a sentence. These are typically classified as function words, serving a grammatical role rather than adding substantial content or meaning.~\cite{jurafsky2000speech,perplexity}. 
Even though function words provide little semantic meaning, not all of them can be effectively reconstructed on the basis of context. Some of them, such as pronouns or quantifiers, cannot be always inferred based on context. For example, in the sentence \textit{"I went to the marathon in the city center"}, if we remove the function words, the sentence will be \textit{"went marathon city center"}. The words "the", "to", and "in" are easily recovered based on the context, but the first pronoun could be any of the personal pronouns.

The categorization and role of function words are clear from the English grammar and linguistic theory. However, few NLP studies have analyzed them regarding how they affect representation learning~\cite{kim2019nlpfunctionwords}. Our interest in this paper lies on how easy it is for a language model to infer the function words based on the context, so then we can select a set of easily-inferable function words to be included in the instruction prompt.
For this purpose, we propose a new algorithm that leverages language models trained with masking technique. This approach enables the sorting of an input word list (i.e. function words, stop words...) based on how easily are them inferred based on the context
The algorithm focuses on calculating the difference in probability between the actual word and the alternative most probable word in a fixed position based on the context. This difference represents the prevalence of the actual word over the distribution of the rest of possible words. This means that, the bigger the difference, the less doubt a language model has about choosing that word over the rest, and, therefore, it is easily-inferable based on the context.

For a  given text $c$ (e.g. a paragraph from a corpus) and a word $w$, let $c'$ be the text $c$ with the word $w$ masked. We define the probability difference as:

\[
\Delta P = P(w \mid c') - P(w_{\rm{alt}} \mid c')
\]

where $P(w_{\rm{alt}} \mid c')$ is the probability of occurrence of the most probable alternative word ($w_{\rm{alt}} \neq w$) given the same context, obtained using the LM. The sorted list of function words based on this criteria is computed following Algorithm \ref{alg:inferable}. After computing $\Delta P$ for all function words in each corpus fragment, it computes the average for each word, and returns the input list of words sorted by this average.

\begin{algorithm}
\caption{Words sorted by ease of inference}
\label{alg:inferable}
\begin{algorithmic}[1]
\small
\State \textbf{Input:} Word set $W = \{w_1, w_2, \dots, w_n\}$,\\
\quad\quad\quad text corpus $C = \{c_1, c_2, \dots, c_m\}$, \\
\quad\quad\quad language model $LM$
\State \textbf{Output:} Sorted W' by probability difference $\Delta P_j$

\For{each $c_i \in C$} 
    \For{each $w_j \in W$}
        \If{$w_j \in c_i$}
            \State Compute $\Delta P_{i,j}$ with $LM$
        \EndIf
    \EndFor
\EndFor

\For{each $w_j \in W$}
    \State $\Delta P_j =$ average$(\Delta P_{i,j})$
\EndFor

\State $W'$ = Sort $W$ by $\Delta P_j$ in descending order
\State \Return W'
\end{algorithmic}
\end{algorithm}

%% file: Sections/5.1.pipeline.tex
\section{Cost-Saving Pipeline} \label{pipeline}
\begin{figure*}[htbp]
    \centering
    \includegraphics[width=0.9\textwidth]{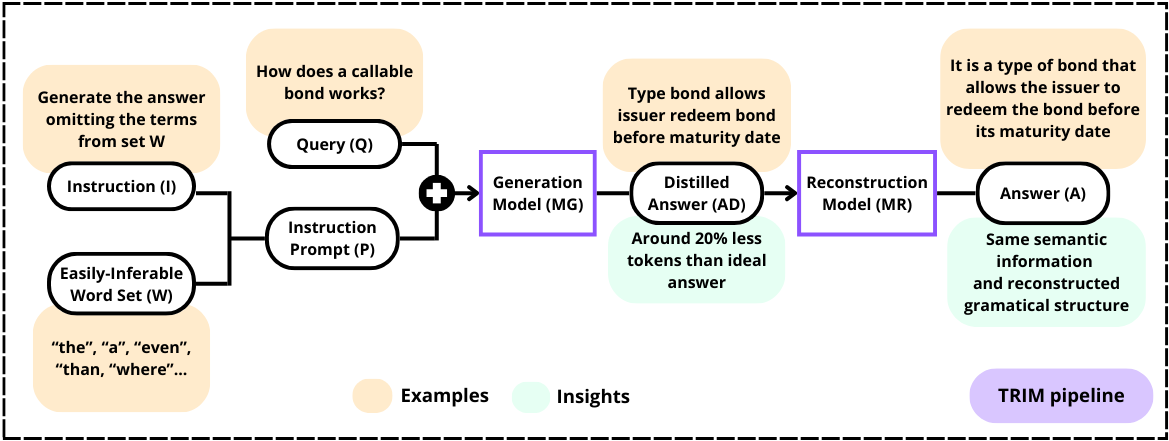}
    \caption{Diagram of the proposed cost-saving pipeline with examples}
    \label{fig:diagram1}
\end{figure*}

The exploratory research showed the feasibility of the initial idea. Now, we cover the development and specification of the pipeline. We propose TRIM (Token Reduction and Inference Modeling pipeline for cost-effective language generation), a pipeline for wrapping the use of generative LLMs (both closed and open) to save costs during output generation of natural language through the philosophy of omitting the generation of easily-inferable low-informative words. Later on, a much smaller and optimized language model handles the reconstruction of the full text. The pipeline, also depicted in Figure~\ref{fig:diagram1}, comprises the following elements:

\textbf{Instruction}: Let $I$ be the instruction where the task of omitting easily-inferable words based on a pre-defined set is explained in detail to the LLM.

\textbf{Easily-Inferable Word Set}: Let $W$ be a set of words with low semantic relevance (i.e. function words, stop words...) deemed as easily-inferable based of the context. This set can be experimentally collected from a corpus using Algorithm~\ref{alg:inferable}.

\textbf{Instruction Prompt}: Let $\mathcal{P}$ be the prompt designed to instruct the LLM in the task of omitting words given $W$ based o the instruction $I$.

\textbf{Generation Model}: Let $M_G$ be the generative LLM  model capable of generating a response to a query, omitting words from $W$. The response is called \textbf{distilled answer} $A_D$.

\textbf{Reconstruction Model}: Let $M_R$ be an encoder-decoder model fine-tuned to reconstruct truncated output into full narrative text with regards of the specific set of words $W$.

Being $Q$ the user query and $A$ the expected answer returned to the user in normal conditions, the pipeline process could be exemplified as: 
\[
A_D = M_G(Q, P(W))
\]
\[
A = M_R(A_D)
\]

%% file: Sections/5.datasets.tex
\section{Datasets \& Models} \label{datasets}
For the experimentation, we used two datasets. Firstly, the Wikipedia dump, which is composed of a large amount of well-written and grammatically correct data on the domain of knowledge~\cite{wikidump}. Different subsets from it will be used for analyzing the ease of inferring functions words,
and also for training and testing reconstruction models.

Secondly, we have created a new evaluation dataset specifically tailored to evaluate the performance of the reconstruction model $M_R$ in the context of the TRIM pipeline where it reconstructs distilled text generated by a given generation model $M_G$, resembling a real use case scenario. In this dataset, called Natural Language Distilled Answers dataset - NaLDA,  we cover five main task types of the natural language generation field: create, knowledge-answer, re-write, summarize and translate. The selection of this 5 task types comes from the research in the literature on natural language generation tasks performed by LLMs, like the building of the Alpaca Dataset~\cite{alpaca}. The NaLDA dataset is composed of 44.800 entries across the 5 task types, where each entry instructs a given generation model $M_G$ to generate, for a given word set $W$, a distilled answer $A_D$ and the ideal answer $A$ to a task. With these two answers, a reconstruction model $M_R$ can be later on evaluated. See supplementary material (Section~\ref{NaLDA-elaboration}) for specifications on the NaLDA dataset elaboration, prompts and examples.

For Algorithm~\ref{alg:inferable}, which generates a sorted set of contextually easily-inferable words, we use the RoBERTa model~\cite{liu2019robertarobustlyoptimizedbert}. RoBERTa efficiently computes word occurrence probabilities at masked positions based on context.

As the generation model $M_G$ for the evaluation, we consider models from (1) OpenAI \cite{openai}, such as GPT-4o (\textit{gpt-4o-2024-08-06}), GPT-4.1-mini (\textit{gpt-4.1-mini-2025-04-14}) and GPT-4.1 (\textit{gpt-4.1-2025-04-14}) and models from (2) Anthropic \cite{anthropic}, including Claude Sonnet~4 (\textit{claude-sonnet-4-20250514}) and Claude Opus 4.1 (\textit{claude-opus-4-1-20250805}).


To evaluate reconstruction models, we trained the following models: (1) BART from Facebook AI, an early transformer with an auto-regressive decoder~\cite{lewis2019bartdenoisingsequencetosequencepretraining}; (2) Google’s T5, known for strong summarization and translation~\cite{t5}; and (3) Alibaba’s Qwen2.5 (0.5B and 1.5B variants), top-ranked among instruct models of similar size according to the Open LLM Leaderboard~\cite{yang2024qwen2technicalreport,open-llm-leaderboard-v2}. These models are much smaller than state-of-the-art LLMs, making them suitable for reconstruction tasks with lower computational costs. Table~\ref{narrators_summary} shows their parameter sizes.

\begin{table}[h!]
\centering
\small
\begin{tabular}{rrrr}
\toprule
 BART & T5 & Qwen2.5-0.5 & Qwen2.5-1.5 \\
\midrule
0.139 & 0.223 & 0.5 & 1.5 \\
\bottomrule
\end{tabular}
\caption{Size in billions of the selected models.}
\label{narrators_summary}
\end{table}

%% file: Sections/7.evaluation.tex
\section{Evaluation}
The evaluation is divided into two experiments. The first one applies  Algorithm~\ref{alg:inferable} for the identification of a good set of easily-inferable words $W$ without notable downgrade in the metrics. In the second experiment, once the set $W$ has been fixed, several language models will be trained and evaluated in the task of the reconstruction model $M_R$. We also analyzed if the results validate the applicability of the complete pipeline without noticing relevant deterioration in the output.

The general evaluation metrics employed to compare texts encompass three primary aspects: grammatical similarity, semantic similarity, and coherence. Grammatical similarity between two texts is assessed using SacreBLEU, METEOR, and ROUGE, which primarily perform n-gram-based comparisons~\cite{post2018sacrebleu, banerjee2005meteor, lin2004rouge}. For semantic similarity, we use cosine similarity, which measures the similarity between two texts using RoBERTa embeddings~\cite{liu2019robertarobustlyoptimizedbert,salton1975cosine}.
For textual coherence, we use the perplexity metric, which quantifies coherence and fluency based on how easily a language model (in our case, GPT-2~\cite{gpt2}) can predict the next tokens, where lower values indicates higher fluency and coherence~\cite{perplexity}.
These experiments were performed on AWS "g4dn.xlarge" machines with Nvidia "T4" GPU with 16GB of memory.

\subsection{Metrics for Reconstruction Task}
Furthermore, we thought it was necessary to find a specific metric that would evaluate the reconstruction of the words from the easily-inferable word set $W$. The general NLP metrics focus on comparing two pieces of text that can come in different sizes (i.e., word, sentence, paragraph...), but they do not offer the versatility of focusing only on a specific set of words according to their exact position in the ideal text, covering the cases of omission, substitution, match or insertion. To this extent, we developed a new evaluation metric specifically focused on evaluating the performance of the reconstruction task, given the ideal text and the reconstructed one. We first had to find a way of matching the two texts, so we can later on check each word position. The Needleman-Wunsch algorithm, used in bio-informatics to align protein or nucleotide sequences, offers this desired feature of alignment~\cite{needleman1970general}. Then, we analyzed and adapted it to our scenario, effectively aligning two texts (one being the original and one being the altered one) even over scenarios where there are missing, different or extra words.

After applying the algorithm, we can compare each  positions in the alignment. Since we know beforehand the set of easily-inferable words $W$ that have been omitted, we can account for scenarios involving omission, substitution, match or insertion for those words. With this information, we obtain precision, recall and F1 values to get a clearer view of the quality of the reconstruction task. We consider as true positives (TP) the match scenario, as false positive (FP) scenarios of inferring a word that is incorrect (insertion and substitution), and as false negative (FN) scenarios where the ideal word is not inferred (omission and substitution). 
From now on, these metrics will be referred as $\Theta$ metrics.

\subsection{Inferable Word Set} \label{inferable_word_set}
In this study, we obtain the set of words $W$ that will be used in the next experiment, analyzing how metrics degrade as the size of this set increases. This will allow us to empirically decide a good choice of $W$. We applied Algorithm~\ref{alg:inferable} using a random sample of 100,000 paragraphs from the Wikipedia dump as the text corpus, an extended list of 127 English function words \cite{jurafsky2000speech} as word set, and RoBERTa as the language model. This returned the list of function words sorted by ease of inference based on the context regarding the text corpus.

To determine the boundary for $W$, we followed an iterative procedure in which the T5 model is fine-tuned for the reconstruction task using the top-N words from the sorted list, increasing the subset in 5 words, from size 5 up to 60 (12 levels). For each of these trainings we used a new random Wikipedia sample containing 100,000 paragraph pairs (\textit{distilled, original}) where the distilled text is built by removing the words of the corresponding level. Additional 5,000 pairs for testing were built with the same schema. The performance is then evaluated at each of these levels, analyzing the degradation of the results as the set size increases. 
 






\begin{figure}[ht]
    \centering
    \begin{subfigure}[b]{\columnwidth}
        \centering
        \includegraphics[width=0.99\textwidth]{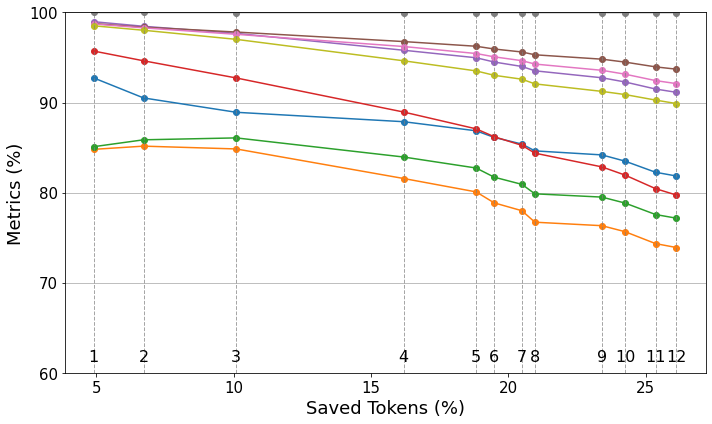}
    \end{subfigure}
    
    
    \begin{subfigure}[b]{\columnwidth}
        \centering
        \includegraphics[width=0.99\textwidth]{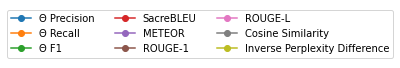}
    \end{subfigure}
    
    \caption{Evaluation at different levels of saved tokens.}
    \label{fig:curve}
\end{figure}

Figure~\ref{fig:curve} shows the metric values plotted against the percentage of saved tokens per level, with trend lines extrapolated. From level 5 onward (about 18\% saved tokens), further increases in saved tokens are minimal while metric degradation continues. Thus, we set the boundary at level 5 and select these 25 function words (set $W$) for the final evaluation.



\subsection{Reconstruction Model}
From this point forward, we use the fixed set $W$ and focus on developing the final reconstruction models $M_R$ to infer the omitted words. For training, we selected a new random sample of 100,000 Wikipedia paragraphs and removed the easily inferable words in $W$, creating the training corpus. The four chosen language models for $M_R$ were fine-tuned on this reconstruction task using the new corpus. Training details are provided in Table~\ref{tab:training_summary} of the Supplementary Material.

We tested each reconstruction model $M_R$ on two evaluation sets. First, on a new sample of 9,000 Wikipedia paragraphs to validate the training, and second, on the NaLDA dataset for each of the generation models $M_G$ (Section~\ref{datasets}).  Table~\ref{metric_llm} presents the NaLDA results with GPT-4o as $M_G$, including metrics comparing the distilled and ideal answers. Detailed results for other generation models are provided in Table~\ref{all_results} of the Supplementary Material.

Regarding the $\Theta$ metrics on the NaLDA dataset (Table~\ref{metric_llm}), overall results indicate good performance (82.8 F1 score or higher), with no significant differences between models. Smaller encoder-decoder models like Bart and T5 show higher precision, while larger decoder-only models such as Qwen2.5 (0.5B and 1.5B) achieve better recall. 
These differences likely stem from their architectures: encoder-decoder models use cross-attention for better context understanding, while decoder-only models tend to be more verbose and infer more terms.

Regarding the rest of metrics, we observe that the grammatical metrics have good values across all models as well, being Qwen2.5(1.5B) the one obtaining the best results. 
All models achieve near-perfect scores on the semantic metric, demonstrating that their outputs effectively preserve the intended meaning.
The perplexity suffers a small increase over the original one, being Qwen2.5(1.5B) the one obtaining the lowest perplexity. These results were obtained with a significant percentage of saved tokens, 18.85\% in the Wikipedia sample and 19.41\% for the answers in the NaLDA dataset. 

\begin{table*}[h!]
\centering
\small
\begin{tabular}{lrrrr|r}
\toprule
 & Bart & T5 & Qwen2.5-0.5 & Qwen2.5-1.5 & Distilled Text\\
\midrule
$\Theta$ Precision (\%) & 88.81(9.59) & \textbf{90.14(9.41)} & 82.11(11.64) & 85.35(10.78) & 95.67(17.31)\\
$\Theta$ Recall (\%) & 82.50(11.04) & 82.24(11.91) & 84.33(10.27) & \textbf{87.65(9.19)} & 8.61(10.50)\\
$\Theta$ F1 (\%) & 85.11(9.36) & 85.51(9.79) & 82.77(9.98) & \textbf{86.10(9.12)} & 14.11(15.51)\\
\midrule
SacreBLEU (\%) & 81.32(9.83) & 82.92(10.69) & 81.59(9.84) & \textbf{83.99(9.25)} & 49.25(12.61)\\
METEOR (\%) & 91.51(5.79) & 92.15(6.85) & 92.73(5.42) & \textbf{93.70(4.84)} & 74.95(7.52)\\
ROUGE-1 (\%) & 94.82(3.33) & 95.13(3.99) & 94.62(3.54) & \textbf{95.26(3.25)} & 85.79(4.47)\\
ROUGE-L(\%)  & 93.56(4.08) & 93.96(4.66) & 93.03(4.43) & \textbf{94.10(3.98)} & 85.78(4.50)\\
Cosine Similarity (\%) & 99.98(0.03) & \textbf{99.99(00.02)} & 99.98(00.02) & 99.98(00.03) & 99.92(00.04)\\
Perplexity & 35.06(18.72) & 31.53(17.46) &  32.52(17.48) & \textbf{31.13(16.54)} & 158.66(139.41)\\
Perplexity Original & 25.59(13.87) & 25.59(13.87) & 25.59(13.87) & 25.59(13.87) & 25.59(13.87)\\
Saved Tokens (\%) & 19.41(5.84) & 19.41(5.84) & 19.41(5.84) & 19.41(5.84) & --- \\
Training Time (h) & 1.5 & 3 & 7.5 & 18 & --- \\
Parameters (B) & 0.139 & 0.223 & 0.5 & 1.5 & --- \\
\bottomrule
\end{tabular}
\caption{Average (standard deviation) metrics results over the NaLDA dataset using GPT-4o as $M_G$}
\label{metric_llm}
\end{table*}

The metrics for the distilled text show significantly lower grammatical scores and higher perplexity, while the semantic metric remains nearly perfect. This indicates that removing the easily-inferable word set $W$ disrupts grammatical structure and coherence but largely preserves the original semantic content.

Since there is no significant difference among models, any can be used effectively for the reconstruction task. Notably, T5 outperforms Qwen2.5 (0.5B) and is comparable to Qwen2.5 (1.5B), even surpassing the latter in some metrics despite being nearly seven times smaller. Given its smaller size, lower cost, and strong performance, T5 appears to be the optimal choice for this task.
Focusing on T5, we compared generation models $M_G$ for generalization (Table \ref{among_llms}). All models perform the task successfully with similar results, despite large differences in price and size. However, comparing models is complicated due to variations in saved tokens, which are inversely related to metric performance. This highlights the value of prior experimentation (Section \ref{inferable_word_set}) in identifying the balance between saved tokens and desired performance, as some models are more verbose and use more function words. Overall, GPT-4o follows specific instructions most effectively, achieving the highest $\theta$ metrics, while Claude models show stronger grammatical metrics but save the fewest tokens.

\begin{table*}[h!]
\centering
\small
\begin{tabular}{lccccc}\toprule
 & GPT-4o & GPT-4.1-mini & GPT-4.1 & Sonnet-4 & Opus-4.1 \\ \midrule
$\Theta$ Precision (\%) & \textbf{90.14(9.41)} & 87.14(11.28) & 88.41(10.10) & 87.15(12.42) & 88.56(9.98) \\
$\Theta$ Recall (\%) &\textbf{ 82.24(11.91)} & 79.64(12.39) & 82.22(12.01) & 81.56(13.74) & 81.12(12.12) \\
$\Theta$ F1 (\%) & \textbf{85.51(9.79)} & 82.52(10.87) & 84.73(10.06) & 83.39(12.51) & 84.03(10.32) \\
\midrule
SacreBLEU (\%) & 82.92(10.69) & 75.21(10.33) & 81.78(9.48) & 85.13(9.00) & \textbf{86.49(8.25)} \\
METEOR (\%) & 92.15(6.85) & 88.49(6.01) & 92.10(5.00) & 93.48(5.26) & \textbf{93.76(5.17)} \\
ROUGE-1 (\%) & 95.13(3.99) & 92.58(3.68) & 94.86(3.09) & 95.71(3.28) & \textbf{96.11(3.09)} \\
ROUGE-L (\%) & 93.96(4.66) & 90.95(4.35) & 93.75(3.70) & 94.81(3.74) & \textbf{95.22(3.50)} \\
Cosine Similarity (\%) & \textbf{99.99(0.02)} & 99.98(0.02) & \textbf{99.99(0.02)} & \textbf{99.99(0.02)} & \textbf{99.99(0.02)} \\
Perplexity & \textbf{31.53(17.46)} & 37.67(20.43) & 35.03(18.20) & 35.36(19.04) & 35.25(18.36) \\
Perplexity Original & 25.59(13.87) & 26.51(16.13) & 27.18(12.89) & 30.85(18.06) & 30.74(15.90) \\
Saved Tokens (\%) & 19.41(5.84) & 22.79(5.44) & 19.45(5.32) & 17.83(5.00) & 17.38(4.36) \\
\bottomrule
\end{tabular}
\caption{Average (standard deviation) metrics results over the NaLDA dataset using T5 as $M_R$}
\label{among_llms}
\end{table*}

We also compared evaluation results between the NaLDA dataset (using GPT-4o as $M_G$) and a Wikipedia sample. Again, the results are consistent with LLM showing the ability to omit only contextually inferable words, rather than all pre-defined words, highlighting its nuanced understanding. See results for Wikipedia sample and discussion in Appendix~\ref{Appendix:validation_result}

\subsubsection{Results across different task types} \label{results_different_task types}

\begin{figure}[ht]
    \centering
    \begin{subfigure}[b]{\columnwidth}
        \centering
        \includegraphics[width=0.99\textwidth]{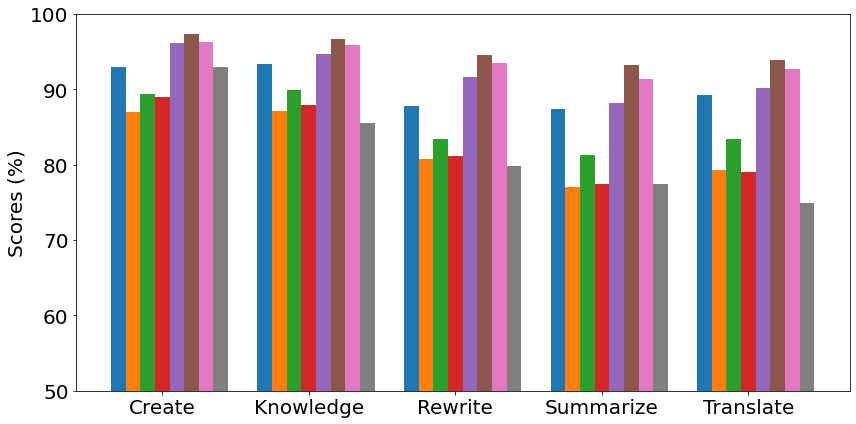}
    \end{subfigure}
    
    
    \begin{subfigure}[b]{\columnwidth}
        \centering
        \includegraphics[width=0.99\textwidth]{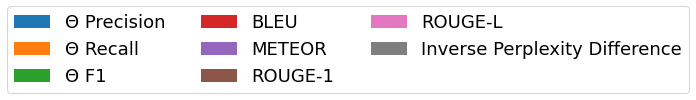}
    \end{subfigure}

    \begin{subfigure}[b]{\columnwidth}
        \centering
        \includegraphics[width=0.99\textwidth]{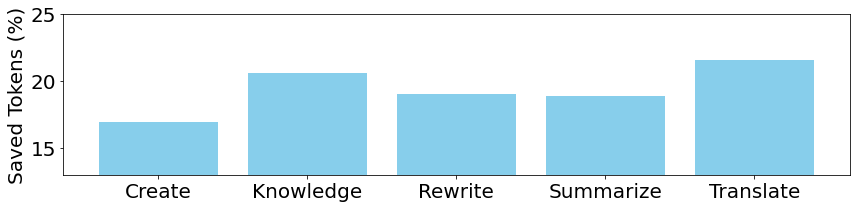}
    \end{subfigure}
    
    \caption{Comparison of metrics for GPT-4o as $M_G$ and T5 as $M_R$ across the 5 different task types of the NaLDA dataset.}
    \label{fig:T5_results}
\end{figure}

Examining results across the five NaLDA task types using GPT-4o as $M_G$ and T5 as $M_R$ (Figure~\ref{fig:T5_results}), several insights emerge. The reconstruction model $M_R$ performs best on the \textit{knowledge-answer} task, where the LLM generates responses to knowledge questions. These outputs are argumentative and align well with the Wikipedia-based training data. In contrast, $M_R$ shows lower performance on rewrite, summarize, and translate tasks, with an average metric decrease of about 5.77\%. This suggests that the TRIM pipeline is more effective when the LLM produces free-form text rather than working from a provided passage.

Additionally, the percentage of saved tokens varies across task types, even though the word set $W$ remains constant. This variation does not directly correlate with evaluation metric performance. For instance, the knowledge-answer task type achieves both high scores and a high percentage of saved tokens, while the create task type saves fewer tokens but still performs well. This suggests that factors such as task type and the nature of the output text also influence token savings. Aside from these differences, all F1 scores are above \~80\%, indicating that the pipeline is effective across all narrative tasks.

%% file: Sections/10.discussion.tex
\section{Discussion}

\subsection{Cost Saving}
Estimating the cost of using LLM API services at token level is a non trivial task~\cite{GPTPricing}.  Costs depend on many factors, including type of task, prompt quality, customization, API call volume, and computational resources. Nevertheless, the model size is the most relevant feature when considering model inference cost. Neither OpenAI nor Anthropic have officially disclosed the size of their most recent models. There have been attempts of estimate the size of these models by comparing them with open-source models, reaching a general consensus estimate for the GPT-4o and Sonnet 3.5 models of around 200 billions parameters, placing models such as Sonnet 4, GPT-4.1 and Opus 4.1 at even higher values~\cite{abacha2025medecbenchmarkmedicalerror, epoch2024frontierlanguagemodelshavebecomemuchsmaller}. Although not knowing their sizes exactly, we want to make an exercise to illustrate how we should think about the potential cost savings of our pipeline. Our proposal should consider the gain/losses of processing text, which for cost purposes is measured by its \textit{size}. The relevant text fragments are:
    
\textbf{Extra inputs ($I$)} : The extra text added to the prompt to specify the task of omitting function words. Specifically for our pipeline, the size of $\mathcal{P}(I,W)$.  

\textbf{Gained output ($G$)}: the gains of saved tokens compared to the vanilla generation. This is the difference between size($A$) and size($A_D$). 

\textbf{Reconstruction inputs ($I_R$)}: the query for the reconstruction task. This is equivalent to size($A_D$).

\textbf{Reconstruction output ($O_R$)}: the text generated for the final output. This is equivalent to size($A$).

On the other hand, we express the input/output costs per size unit using the same subscripts as models. So, let $C^i_D, C^o_D$ be the input and output cost for the generation model, and $C^i_R, C^o_R$ the input and output cost for the reconstruction model. The requirement framework for which our pipeline application would make sense is expressed by the formula: 
\[
C^o_D G \geq C^i_D I + C^i_R I_R + C^o_R O_R 
\]

In our evaluation, reconstruction models range from 0.139 to 1.5 billion parameters, which is at least two orders of magnitude smaller than the size estimation of the generation models $M_G$ employed like GPT-4o, GPT-4.1, Sonnet 4 and Opus 4.1.
Therefore, the relevant term in the right hand side is $C^i_D I$. Using available prices for the GPT and Claude API services as a proxy of cost, we see that the ratio $C^i_D/C^o_D$ is $\sim$0.25 for GPT and $\sim$0.2 for Claude.  In our experiments, instruction $I$ adds 169 extra tokens to the prompt (See prompt 2 in Supplementary Material \ref{supplementary_exploratory}), so it would be compensated after 42 and 34 omitted tokens (fewer number of words) for the GPT and Claude models respectively. The final results depends on the tokens saved by each model using TRIM, but to exemplify with GPT-4o (19.4\% saved tokens) and Sonnet 4 (17.83\% saved tokens), natural language generation answers longer than $\sim$216 tokens and $\sim$190 tokens will have better inference cost with our pipeline respectively.

\subsection{Other Language Opportunities}
Although this study focused on English, the technique is language independent. Most languages follow the Zipf distribution~\cite{piantadosi2014zipf}, and frequency ranks are highly correlated across languages~\cite{calude2011we}, though prior analyses emphasized word semantics over function words. Our final easily-inferable word set $W$ overlaps with only four words from the Swadesh list~\cite{swadesh1952lexico}, leaving language-specific benefits as an open question for future work. Languages like Spanish, French, and Italian, which rely on frequent function words for grammatical cohesion, may adapt well to this approach, while languages with greater morphological complexity, such as Finnish or Turkish, may present additional challenges.


%% file: Sections/11.conclusion.tex
\section{Conclusion}


We proposed TRIM, a pipeline for reducing LLM text generation costs by omitting words that are easily inferred and reconstructing them with a smaller model trained for this task. To support this, we introduced an algorithm for ranking word inferability, built the NaLDA evaluation dataset across five generation tasks, and presented a new reconstruction metric. Our experiments with four reconstruction models showed strong performance and an average token savings of 19.41\% for GPT-4o, demonstrating TRIM’s potential for improving cost efficiency in natural language generation.

%% file: Sections/13.limitations.tex
\section{Limitations}
\label{limitations}
We acknowledge some limitation this pipeline could have both during evaluation and in possible application scenarios out of scope. In the evaluation, we have attempted to cover a wide range of perspectives, from semantic content to grammatical structure evaluations, including the coherence of the text and a new metric specifically focused on the task of text reconstruction, paying attention to the subset of terms expected to be omitted and reconstructed. Thus, there are scenarios where automated evaluation may error and excessively penalize the effectiveness of the entire pipeline.
Examples of this occurrence can happen in scenarios where function words in the text can be omitted or exchange with others while the sentence remains grammatically and semantically correct, preserving also its coherence. An example can be the sentence "I work from 9 to 5." If we compare it with the sentence "I work 9 till 5," the function word "from" has been omitted and the function word "to" has been replaced by "till." While the evaluation of semantic meaning will indicate that these two sentences are equivalent, the rest of the general and specific grammatical structure metrics will indicate lower performance. Therefore, a human evaluation of the entire pipeline would be interesting to assess these types of scenarios as well.

Although the experiments explored the capability of an LLM to generate output without including words that could be inferred later, it did not consider how this might negatively impact the LLM's performance. LLMs are trained on vast datasets of human-written text that follow grammatical rules. Consequently, their output is based on this training data and on the sequence of tokens already produced during a session. If the usual grammatical structure is absent at generation and at the previous generated data, this could disrupt the ongoing token generation. For example, a previous study remarks the importance of a (different) set of tokens at generation time called \textit{stylistic tokens} (e.g., ‘Hello’, ‘Thank’, ‘However’, ‘Remember’, etc.), which include transitional phrases, discourse markers, and safety disclaimers~\cite{lin2023unlockingspellbasellms}. Nevertheless, in a proper implementation, the potential penalty would only affect new generation of text. After generating a piece of text, the reconstruction model would restore its grammatical structure. The distilled piece of text in the context window would then be replaced by the reconstructed one. For subsequent generations, as the intrinsic grammatical structure was restored, the LLM would not face any penalties regarding previous generations.

In this study we have primarily focused on the natural language generation domain, where structured and verbose content enables effective token-saving by omitting low-information words with minimal impact on semantic clarity. However, generalizing this pipeline to other domains is not straightforward, like conversational as it could lack of enough amount of easily-inferable low-semantic words for the trade-off of adding extra tokens at the initial instruction, or code generation where one bad inferred word could imply an execution error. Moreover, during experimentation neither the so called "reasoning" models like o3 nor the reasoning capabilities of models like Clause Opus 4.1 or Sonnet 4 were used. Despite believing that the reasoning process can greatly benefit from the use of this pipeline from the cost perspective, the focus of this research has been on the response generated by the model itself and its applicability for the reasoning step is left for future research.

%% file: Sections/12.disclaimer.tex
\section*{Disclaimer}
This paper was prepared for informational purposes by the Artificial Intelligence Research group of JPMorgan Chase \& Co. and its affiliates ("JP Morgan'') and is not a product of the Research Department of JP Morgan. JP Morgan makes no representation and warranty whatsoever and disclaims all liability, for the completeness, accuracy or reliability of the information contained herein. This document is not intended as investment research or investment advice, or a recommendation, offer or solicitation for the purchase or sale of any security, financial instrument, financial product or service, or to be used in any way for evaluating the merits of participating in any transaction, and shall not constitute a solicitation under any jurisdiction or to any person, if such solicitation under such jurisdiction or to such person would be unlawful.

© 2025 JPMorgan Chase \& Co. All rights reserved

%% file: Sections/14.Supplementary_Material.tex
\appendix
\clearpage
\section{Supplementary Material}

\subsection{Capabilities of LLMS on Omitting Function Words} \label{supplementary_exploratory}
For this exploratory experiment we have used the model "GPT-4o" OpenAI through its API in its version "gpt-4o-2024-08-06". The list of articles and conjunctions used are: \textit{the, a, an, as, though, because, before, even, if, that, since, so, than, unless, until, when, whenever, where, whereas, wherever, while, and, but}. The set of prompts used in this experimentation are the following:

Prompt 1: \textit{Respond just in a paragraph. $\{query\}$}

Prompt 2: \textit{Respond just in a paragraph in a distilled way removing the redundant and semantic-irrelevant words of the language such as: $\{list\_of\_words\}$. That is to say, omit the words of the ideal answer that do not affect directly to the semantic meaning and can be reconstructed later on based on the context. This task will be carried out by another language model in charge of generating those omitted words based on the list provided and reconstructing the ideal answer. Return only the distilled answer, nothing else. $\{query\}$}

\begin{figure}[h]
    \centering
    \includegraphics[width=\linewidth]{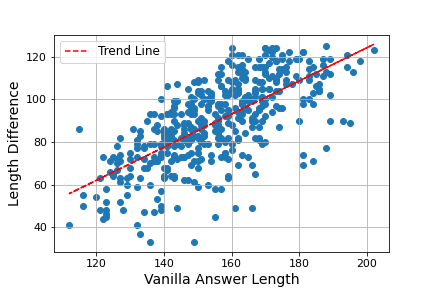}
    \caption{Vanilla Answer Length vs Length Difference Between Answers}
    \label{fig:scatter_plot}
\end{figure}

Figure~\ref{fig:scatter_plot} shows the relationship between the length (measured in tokens) of the vanilla answer (answer to prompt 1) and the difference in tokens between the vanilla answer and distilled answer (answer to prompt 2), showing how the trend line grows linearly. It is important to note that both answers were generated in separate sessions, so their differences cannot be fully explained simply by omitted tokens.

\subsection{NaLDA Dataset Elaboration} \label{NaLDA-elaboration}
To build this dataset, we used two approaches depending on the task type. For knowledge question-answering (QA) and create tasks, we generated synthetic questions using various templates for diversity. For knowledge QA tasks, we created 9,000 questions across 30 general topics (such as Science, History, Art). For each topic, we selected 20 concepts and 5 keywords per concept. We then used templates to generate questions focused on explanations and comparisons between keywords and concepts. The topics, concepts, and keywords were generated with GPT-4o and reviewed by humans for accuracy. Below is an example of a template question using the keyword "decentralization" linked to the concept "blockchain", related to one of the topics "technology".

\begin{itemize}
    \item What is decentralization in the context of blockchain?
    \item Who are the main contributors of decentralization in the context of blockchain and a summary of their contribution?
    \item Define in what techniques or research has decentralization influenced and how, out of the context of blockchain
    \item What is the state of the art and future or expected development of decentralization in the context of blockchain?
    \item Summarize the concepts relevant for the general knowledge and understanding of decentralization in the context of blockchain?
    \item What is the difference between decentralization and smart contracts in the context of blockchain?
\end{itemize}

For the create tasks, we have gathered 8,800 different instructions related to create a story. Each story creation task is composed of a place, a theme and a historical character from a sample of 20 places, 20 themes and 20 historical characters. These sets have been also generated by the LLM with a human curation after generation to verify its validity. Here is a set of examples of tasks generated:

\begin{itemize}
    \item Write a story located in a Pirate Ship, with Abraham Lincoln as the main character and a theme of eroism and Sacrifice.
    \item Imagine a tale set in a Industrial Revolution Factory, featuring Napoleon Bonaparte and exploring the theme of Survival and Resilience.
    \item Create a narrative in a Underwater City, where Leonardo da Vinci faces challenges related to Family and Legacy.
    \item Develop a plot in Santiago Bernabeu, starring Albert Einstein and centered around Fantasy and Magic.
    \item Craft a story in a Arctic Base, with Cristiano Ronaldo navigating the complexities of Mystery and Intrigue.
\end{itemize}

For the re-write, summarize, and translate tasks, we sampled paragraphs from the Wikipedia text corpus and created instructions for each task type, resulting in 9,000 entries per type. For re-write tasks, the LLM was told to change the style. For summarize tasks, we used longer paragraphs to better fit the task and instructed the LLM to summarize them. For translate tasks, we selected Spanish Wikipedia paragraphs and asked the LLM to translate them into English. In our pipeline, each of these entries corresponds to query $Q$.

Furthermore, we have to amend each query with the custom prompt $P'$ that instructs the generator model $M_G$ to generate both the ideal and distilled version of the answer. This custom prompt $P'$ is a variation of the instruction prompt $P$ of the TRIM pipeline where, in addition to generate the distilled answer based on the given set of words $W$, it has to generate also the ideal answer so we can evaluate later on. Ideally, the generation of the ideal and distilled answer should be requested in different sessions, so the distilled one would simulate the real scenario of our pipeline, and the ideal would be the ground truth. 
Otherwise, the LLM would have the context of the ideal to generate the distilled version. 
However, after carrying some experiments we realized that doing it in different sessions produce additional variability (e.g., synonyms, grammatical alternative and paraphrasing) that affects directly the evaluation purpose of this dataset, preventing us of using them in the evaluation (see Supplementary Material~\ref{consistency}).
Under the assumption that the LLM can generate distilled text directly as shown in Section~\ref{exploratory}, we keep the single session generation to focus the evaluation on the ability of reconstructing the distilled narrative.

Here is the custom prompt $P'$ that is used together with each task to form the dataset: 
\textit{
I want you to return only a json with two fields ``original\_text'' and ``distilled\_text''. The original text is the response to the question in one paragraph. Then, for the distilled text, return the same response but removing the redundant terms of the language such as: $\{\text{set\_of\_easily-inferable\_words}\}$ that do not affect directly to the semantic and can be reconstructed later on based on the context. That is to say, omit the words of the ideal answer that do not affect directly to the semantic meaning and can be reconstructed later on based on the context. This task will be carried out by another language model in charge of generating those omitted words based on the list provided and reconstructing the ideal answer. Keep the rest of the terms and the structure of the original response, only remove those redundant terms. Here is the task: $\{\text{query}\}$
}

\subsection{Consistency Among Responses in LLMs} \label{consistency}
We have also carried out an experimentation on the consistency among different LLM answers which conditions the evaluation of this paper when dealing with generative models. We have taken a set of 20 knowledge questions, getting the answer of each of the questions 10 times in different sessions (with temperature 0 looking for maximum replicability), and compared the successive generations with the first one. As can be seen in the Table~\ref{tab:consistency}, although the semantic content measured by cosine similarity gives a perfect result, the other metrics related to n-grams and grammatical structure are weaker. This indicates that among the responses the semantic content is maintained, but their grammatical structure varies considerably.

\begin{table}[h!]
\centering
\small
\begin{tabular}{lrr}
\toprule
{} &  Mean &  Variance \\
\midrule
BLEU              &  0.77 &      0.03 \\
ROUGE-1           &  0.87 &      0.01 \\
ROUGE-L           &  0.83 &      0.02 \\
METEOR            &  0.83 &      0.02 \\
Cosine Similarity &  1.00 &      0.00 \\
\bottomrule
\end{tabular}
\caption{Exploratory results on the consistency among answers to the same questions by LLM.}
\label{tab:consistency}
\end{table}

\subsection{Aligning Function Words and $\Theta$-metrics}
$\Theta$-metrics are the metrics designed and used in our evaluation with the purpose of considering only the performance over the function words prediction or reconstruction. Consider the following example:
\begin{itemize}
    \item Original: \textit{The history of art is the fascinating subject of human culture}.
    \item Reconstructed:\textit{ History of the art is a fascinating subject of human culture.}
\end{itemize}
After removing all non-function words and aligning the resulting sequences with the Needleman-Wunsch algorithm \cite{needleman1970general}, we can compute matches and errors as in Table~\ref{tab:matchseq}. In terms of
binary classification result, the matches correspond to the true positives. The false positives are the insertions plus the substitutions. The false negative are the omission plus the substitutions. Then, precision, recall and F1 score are computed in the standard way.

\begin{table*}[htb]
\centering
\small
\begin{tabular}{lcccccc}
\toprule
Original &  The & of & --- & the & of \\
Predict & --- & of & the & a & of \\
\midrule
Result & omission & match & insertion & substitution & match \\ 
\bottomrule
\end{tabular}
\caption{Example of aligned function words and matching result.}
\label{tab:matchseq}
\end{table*}

\begin{table*}[hbt]
\centering
\small
\begin{tabular}{lrrrr}
\toprule
 & BART & T5 & Qwen2-0.5 & Qwen2-1.5 \\
\midrule
Training size & 100.000 & 100.000 & 100.000 & 100.000 \\
Epochs & 5 & 5 & 3 & 2.5 \\
Learning Rate & 5e-5 & 5e-5 & 4e-5 & 3e-5 \\
Batch Size & 8 & 8 & 4 & 4 \\
Gradient step & 8 & 16 & 32 & 8 \\
Weight decay & 0.01 & 0.01 & 0.1 & 0.1 \\
FP16 & True & True & True & True \\
\midrule
Training (hours) & 1.5 & 3 & 7.5 & 17 \\
\bottomrule
\end{tabular}
\caption{Summary of the training process for the selected models.}
\label{tab:training_summary}
\end{table*}


\subsection{Results on Validation Set}
\label{Appendix:validation_result}
Looking at the differences between evaluation sets (NaLDA dataset using GPT-4o as $M_G$ and Wikipedia sample), we observe that the results of the $\Theta$ metrics are better on the NaLDA dataset, while the rest of the metrics are slightly better on average in the Wikipedia sample (See Table~\ref{metric_split}). This indicates that, while the reconstruction model generally performs better on the validation dataset—built with the same philosophy as the training data—it achieves better results on metrics focused solely on reconstructing the pre-defined word set when the LLM chooses which terms to omit during generation, rather than omitting all of them directly.

Furthermore, an interesting insight emerges in the $\Theta$ metrics of the distilled text (i.e., matching distilled text with the ground truth). In the Wikipedia sample evaluation, the $\Theta$-recall is 0 because these words are removed directly from the text. In contrast, in the NaLDA dataset evaluation, the $\Theta$-recall is $8.6\%$, showing that the LLM does not erase all words from the list, but only those it considers easily-inferable from context, as instructed in the prompt (see Appendix  \ref{NaLDA-elaboration}).

Both of these insights demonstrate a desirable capability of the LLM: not only omitting a pre-defined set of words during generation, but also identifying when a word is not easily-inferable based on context and choosing not to omit it, even if it is in the pre-defined word set $W$. These results are consistent with the exploratory analysis shown in Table~\ref{tab:exploratory}.

\begin{table*}[hbt]
\centering
\small
\begin{tabular}{lrrrr|r}
\toprule
 & Bart & T5 & Qwen2.5-0.5 & Qwen2.5-1.5 & Distilled Text\\
\midrule
$\Theta$ Precision (\%) & 88.44(12.67) & \textbf{88.50(12.64)} & 80.30(14.96) & 83.60(14.07) & 100.00(00.00)\\
$\Theta$ Recall (\%) & 81.58(14.74) & 81.61(14.93) & 82.23(14.42) & \textbf{86.14(13.02)} & 00.00(00.00)\\
$\Theta$ F1 (\%) & 84.33(12.71) & 84.39(12.85) & 80.73(13.62) & \textbf{84.40(12.57)} & 00.00(00.00)\\
\midrule
SacreBLEU (\%) & 87.02(8.77) & 88.38(8.53) & 86.01(9.06) & \textbf{88.51(8.28)} & 51.00(13.22)\\
METEOR (\%) & 94.87(4.57) & 95.55(4.19) & 95.77(3.78) & \textbf{96.61(3.30)} & 76.10(7.11)\\
ROUGE-1 (\%) & 96.61(2.89) & \textbf{96.92(2.77)} & 96.19(3.10) & 96.79(2.85) & 86.19(4.38)\\
ROUGE-L (\%) & 95.81(3.46) & \textbf{96.14(3.30)} & 94.99(3.87) & 95.94(3.46) & 86.19(4.38)\\
Cosine Similarity (\%) & \textbf{99.99(00.04)} & \textbf{99.99(00.03)} & 99.98(00.03) & 99.98(00.04) & 99.92(00.07)\\
Perplexity & 42.91(31.48) & 39.35(29.46) & 40.52(29.48) & \textbf{39.01(27.76)} & 208.21(200.66)\\
Perplexity Original & 37.28(26.13) & 37.28(26.13) & 37.28(26.13) & 37.28(26.13) & 37.28(26.13)\\
Saved Tokens (\%) & 18.85(5.87) & 18.85(5.87) & 18.85(5.87) & 18.85(5.87) & --- \\
Training Time (h) & 1.5 & 3 & 7.5 & 18 & --- \\
Parameters (B) & 0.139 & 0.223 & 0.5 & 1.5 & --- \\
\bottomrule
\end{tabular}
\caption{Average (standard deviation) metric results over the evaluation dataset from the Wikipedia dump.}
\label{metric_split}
\end{table*}

\begin{figure*}[!htp]
    \centering
    \begin{subfigure}[b]{0.65\textwidth}
        \centering
        \includegraphics[width=0.6\textwidth]{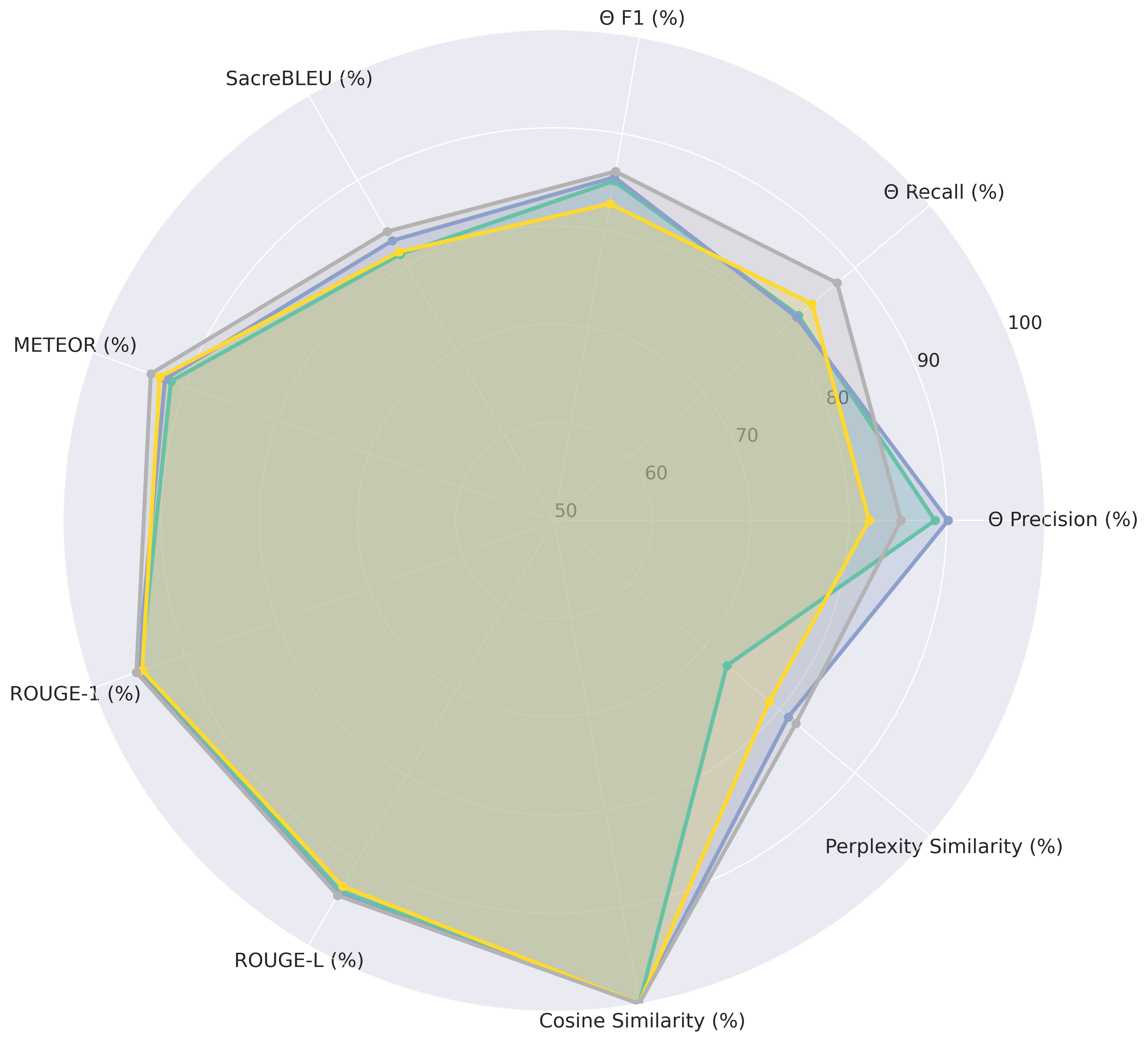}
    \end{subfigure}
    \begin{subfigure}[b]{0.88\textwidth}
        \centering
        \includegraphics[width=\textwidth]{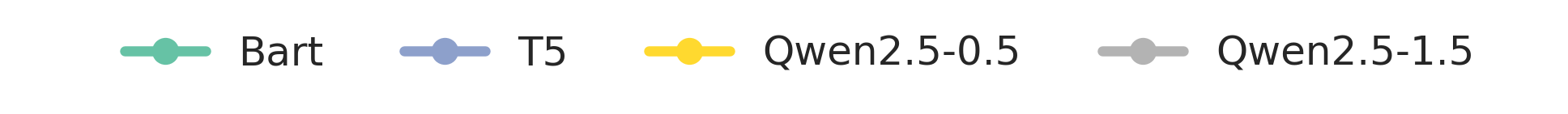}
    \end{subfigure}
    \caption{Visual comparison of the results for the reconstruction models in the NaLDA dataset using GPT-4o as $M_G$}
    \label{fig:radar}
\end{figure*}

\begin{table*}[h!]
\centering
\small
\begin{tabular}{lcccc}\toprule
Metric & Bart & T5 & Qwen2.5-0.5 & Qwen2.5-1.5 \\ \midrule
\multicolumn{5}{c}{\textbf{GPT-4o}} \\ \midrule
$\Theta$ Precision (\%) & 88.81(9.59) & 90.14(9.41) & 82.11(11.64) & 85.35(10.78) \\
$\Theta$ Recall (\%) & 82.50(11.04) & 82.24(11.91) & 84.33(10.27) & 87.65(9.19) \\
$\Theta$ F1 (\%) & 85.11(9.37) & 85.51(9.79) & 82.77(9.98) & 86.10(9.12) \\
\midrule
SacreBLEU (\%) & 81.32(9.83) & 82.92(10.69) & 81.59(9.84) & 83.99(9.25) \\
METEOR (\%) & 91.51(5.79) & 92.15(6.85) & 92.73(5.42) & 93.70(4.84) \\
ROUGE-1 (\%) & 94.82(3.33) & 95.13(3.99) & 94.62(3.54) & 95.26(3.25) \\
ROUGE-L (\%) & 93.56(4.08) & 93.96(4.66) & 93.03(4.43) & 94.10(3.98) \\
Cosine Similarity (\%) & 99.98(0.03) & 99.99(0.02) & 99.98(0.02) & 99.98(0.03) \\
Perplexity & 35.06(18.72) & 31.53(17.46) & 32.52(17.48) & 31.13(16.54) \\
Perplexity Original & 25.59(13.87) & 25.59(13.87) & 25.58(13.84) & 25.58(13.85) \\
Saved Tokens (\%) & 19.41(5.84) & 19.41(5.84) & 19.41(5.83) & 19.41(5.83) \\
\midrule
\multicolumn{5}{c}{\textbf{GPT-4.1-mini}} \\ \midrule
$\Theta$ Precision (\%) & 86.29(11.35) & 87.14(11.28) & 78.61(13.06) & 82.13(12.27) \\
$\Theta$ Recall (\%) & 79.79(11.95) & 79.64(12.39) & 81.98(11.23) & 85.60(10.29) \\
$\Theta$ F1 (\%) & 82.28(10.60) & 82.52(10.87) & 79.64(11.08) & 83.25(10.31) \\
\midrule
SacreBLEU (\%) & 73.16(10.02) & 75.21(10.33) & 73.29(10.01) & 76.02(9.64) \\
METEOR (\%) & 87.59(5.73) & 88.49(6.01) & 88.68(5.50) & 89.85(5.13) \\
ROUGE-1 (\%) & 92.07(3.42) & 92.58(3.68) & 91.60(3.65) & 92.31(3.50) \\
ROUGE-L (\%) & 90.30(4.16) & 90.95(4.35) & 89.45(4.51) & 90.72(4.20) \\
Cosine Similarity (\%) & 99.98(0.03) & 99.98(0.02) & 99.97(0.03) & 99.97(0.03) \\
Perplexity & 42.06(21.90) & 37.67(20.43) & 39.30(20.42) & 37.05(19.42) \\
Perplexity Original & 26.52(16.33) & 26.51(16.13) & 26.40(15.46) & 26.35(14.81) \\
Saved Tokens (\%) & 22.80(5.44) & 22.79(5.44) & 22.80(5.43) & 22.81(5.42) \\
\midrule
\multicolumn{5}{c}{\textbf{GPT-4.1}} \\ \midrule
$\Theta$ Precision (\%) & 87.73(10.19) & 88.41(10.10) & 79.53(12.38) & 83.20(11.51) \\
$\Theta$ Recall (\%) & 82.31(11.65) & 82.22(12.01) & 84.09(11.00) & 87.51(9.91) \\
$\Theta$ F1 (\%) & 84.46(9.79) & 84.73(10.06) & 81.23(10.56) & 84.84(9.65) \\
\midrule
SacreBLEU (\%) & 79.66(9.57) & 81.78(9.48) & 79.23(9.60) & 81.60(9.17) \\
METEOR (\%) & 91.11(5.14) & 92.10(5.00) & 92.07(4.89) & 92.93(4.58) \\
ROUGE-1 (\%) & 94.30(3.15) & 94.86(3.09) & 93.79(3.39) & 94.42(3.25) \\
ROUGE-L (\%) & 93.09(3.80) & 93.75(3.70) & 92.28(4.16) & 93.31(3.88) \\
Cosine Similarity (\%) & 99.98(0.02) & 99.99(0.02) & 99.98(0.02) & 99.98(0.03) \\
Perplexity & 38.84(19.61) & 35.03(18.20) & 36.43(18.41) & 34.96(17.52) \\
Perplexity Original & 27.18(12.89) & 27.18(12.89) & 27.18(12.89) & 27.18(12.89) \\
Saved Tokens (\%) & 19.45(5.32) & 19.45(5.32) & 19.45(5.32) & 19.46(5.32) \\
\midrule
\multicolumn{5}{c}{\textbf{Sonnet-4}} \\ \midrule
$\Theta$ Precision (\%) & 86.46(12.59) & 87.15(12.42) & 77.68(14.17) & 81.53(13.39) \\
$\Theta$ Recall (\%) & 82.75(12.97) & 81.56(13.74) & 84.01(12.53) & 87.82(11.44) \\
$\Theta$ F1 (\%) & 83.73(12.09) & 83.39(12.51) & 79.79(12.66) & 83.70(11.88) \\
\midrule
SacreBLEU (\%) & 84.36(8.43) & 85.13(9.00) & 83.56(8.53) & 86.25(7.84) \\
METEOR (\%) & 93.61(4.36) & 93.48(5.26) & 94.48(4.15) & 95.43(3.55) \\
ROUGE-1 (\%) & 95.73(2.72) & 95.71(3.28) & 95.33(2.95) & 96.01(2.67) \\
ROUGE-L (\%) & 94.81(3.23) & 94.81(3.74) & 93.95(3.64) & 95.11(3.16) \\
Cosine Similarity (\%) & 99.99(0.03) & 99.99(0.02) & 99.98(0.02) & 99.98(0.02) \\
Perplexity & 38.04(19.73) & 35.36(19.04) & 35.86(18.54) & 36.52(223.85) \\
Perplexity Original & 30.84(18.05) & 30.85(18.06) & 30.79(17.60) & 30.81(17.83) \\
Saved Tokens (\%) & 17.83(5.00) & 17.83(5.00) & 17.83(4.99) & 17.83(4.99) \\
\midrule
\multicolumn{5}{c}{\textbf{Opus-4.1}} \\ \midrule
$\Theta$ Precision (\%) & 88.17(10.03) & 88.56(9.98) & 79.26(12.32) & 83.34(11.37) \\
$\Theta$ Recall (\%) & 83.19(11.23) & 81.12(12.12) & 84.53(10.76) & 88.56(9.41) \\
$\Theta$ F1 (\%) & 85.05(9.75) & 84.03(10.32) & 81.15(10.73) & 85.27(9.71) \\
\midrule
SacreBLEU (\%) & 86.56(7.35) & 86.49(8.25) & 85.85(7.43) & 88.61(6.69) \\
METEOR (\%) & 94.51(4.01) & 93.76(5.17) & 95.59(3.68) & 96.54(3.01) \\
ROUGE-1 (\%) & 96.50(2.40) & 96.11(3.09) & 96.26(2.60) & 96.94(2.24) \\
ROUGE-L (\%) & 95.62(2.86) & 95.22(3.50) & 94.89(3.25) & 96.06(2.73) \\
Cosine Similarity (\%) & 99.98(0.03) & 99.99(0.02) & 99.98(0.02) & 99.98(0.02) \\
Perplexity & 37.48(18.85) & 35.25(18.36) & 35.28(17.95) & 33.78(17.07) \\
Perplexity Original & 30.74(15.90) & 30.74(15.90) & 30.75(15.90) & 30.75(15.89) \\
Saved Tokens (\%) & 17.38(4.36) & 17.38(4.36) & 17.38(4.36) & 17.38(4.37) \\
\bottomrule
\end{tabular}
\caption{Average (standard deviation) metrics results over the NaLDA dataset for each $M_G$ and $MR$ models using the word set $W$ obtained during evaluation.}
\label{all_results}
\end{table*}